\documentclass[runningheads]{llncs}
\usepackage{colortbl} 
\usepackage{makecell} 
\usepackage{authblk}
\usepackage{multirow, multicol}
\usepackage[T1]{fontenc}
\usepackage{marvosym} 
\usepackage{amsmath, amssymb, bm}
\usepackage{graphicx}
\usepackage[dvipsnames]{xcolor}

\begin{document}
\title{Toward Patient-specific Partial Point Cloud to Surface Completion for Pre- to Intra-operative Registration in Image-guided Liver Interventions
}
\titlerunning{Surface Completion for Pre- to Intra-operative Liver Registration}

\author{Nakul Poudel\inst{1}\textsuperscript{(\Letter)} \and
Zixin Yang\inst{1} \and
Kelly Merrell\inst{1} \and Richard Simon \inst{2} \and Cristian A. Linte\inst{1,2}}

\authorrunning{N. Poudel {\it et al}.}

\institute{Center for Imaging Science, Rochester Institute of Technology, Rochester, NY 14623, USA \\
\email{np1140@rit.edu}
\and
Biomedical Engineering, Rochester Institute of Technology, Rochester, NY 14623, USA}
\maketitle              
\begin{abstract}

Intra-operative data captured during image-guided surgery lacks sub-surface information, where key regions of interest, such as vessels and tumors, reside. Image-to-physical registration enables the fusion of pre-operative information and intra-operative data, typically represented as a point cloud. However, this registration process struggles due to partial visibility of the intra-operative point cloud. In this research, we propose a patient-specific point cloud completion approach to assist with the registration process. Specifically, we leverage VN-OccNet to generate a complete liver surface from a partial intra-operative point cloud. The network is trained in a patient-specific manner, where simulated deformations from the pre-operative model are used to train the model. 
First, we conduct an in-depth analysis of VN-OccNet’s rotation-equivariant property and its effectiveness in recovering complete surfaces from partial intra-operative surfaces. Next, we integrate the completed intra-operative surface into the Go-ICP registration algorithm to demonstrate its utility in improving initial rigid registration outcomes. Our results highlight the promise of this patient-specific completion approach in mitigating the challenges posed by partial intra-operative visibility. The rotation equivariant and surface generation capabilities of VN-OccNet hold strong promise for developing robust registration frameworks for variations of the intra-operative point cloud.

\end{abstract}

\section{Introduction}

Registration methods play a vital role during image-guided interventions in assisting surgeons to target the key regions of interest, such as tumors and vessels, that lie beneath the organ surface. Image-to-physical registration aligns pre-operative Computed Tomography (CT) or Magnetic Resonance Imaging (MRI) data, represented as a point cloud or mesh, to intra-operative data acquired using surgical tracking devices or cameras, often represented as a point cloud \cite{heiselman2024image,yang2023learning}.  However, the registration \cite{yang2023learning,heiselman2018characterization} faces challenges caused by the partial intra-operative visibility that arises due to constrained camera viewpoints and occlusions~\cite{heiselman2018characterization,jia2021improving,lin2016video}. Therefore, it is necessary to address the issue of partial visibility, which prevents registration methods from performing with sufficient accuracy.

The reconstruction of an intra-operative 3D liver surface from a partially visible surface has the potential to mitigate the partial visibility issue faced by registration methods. Toward this effort, Jia {\it \textit{et al.}} \cite{jia2021improving} proposed a non-rigid registration framework that integrates a learning-based point completion network to generate a complete surface from sparse intra-operative data to guide non-rigid registration. However, the proposed method still requires a rigid registration as the initialization. Foti {\it et al}.~\cite {foti2020intraoperative} adopted a Variational Autoencoder on pre-operative models to generate a full liver surface from a partial point cloud, which undergoes an iterative optimization procedure to generate an intra-operative surface. However, this approach also requires manually identified correspondences between the generated mesh and the intra-operative point cloud, which is non-trivial.

In this work, we propose a patient-specific pipeline to improve image-to-physical registration (and hence pre- to intra-operative registration) by completing partial intra-operative liver surfaces. The pipeline leverages a vector neuron-based occupancy network (VN-OccNet~\cite{deng2021vector}) to recover a complete liver mesh from a partial intra-operative point cloud. VN-OccNet offers two key advantages for the registration: (1) rotation equivariance, which addresses the failure of conventional models under varying orientations of intra-operative data~\cite{poudel2025evaluation}, and (2) the ability to generate watertight meshes rather than point clouds, enabling uniform surface sampling—a crucial feature for registration methods requiring consistent point density~\cite{yang2015go,yang2023learning}. We adopt a patient-specific training strategy that synthesizes intra-operative liver surfaces by deforming a pre-operative patient-specific liver model, hence allowing the model to focus on patient-specific geometry and deformation patterns. We first evaluate VN-OccNet’s ability to generate rotation-equivariant surfaces across diverse rotation settings. We then compare registration outcomes using completed surfaces against those using only partial point clouds. Results show that the reconstructed surfaces significantly reduce registration error, highlighting the benefit of surface completion towards enhancing registration accuracy.

\section{Methodology}

\begin{figure}[!ht]
    \centering
    \includegraphics[width=\textwidth]{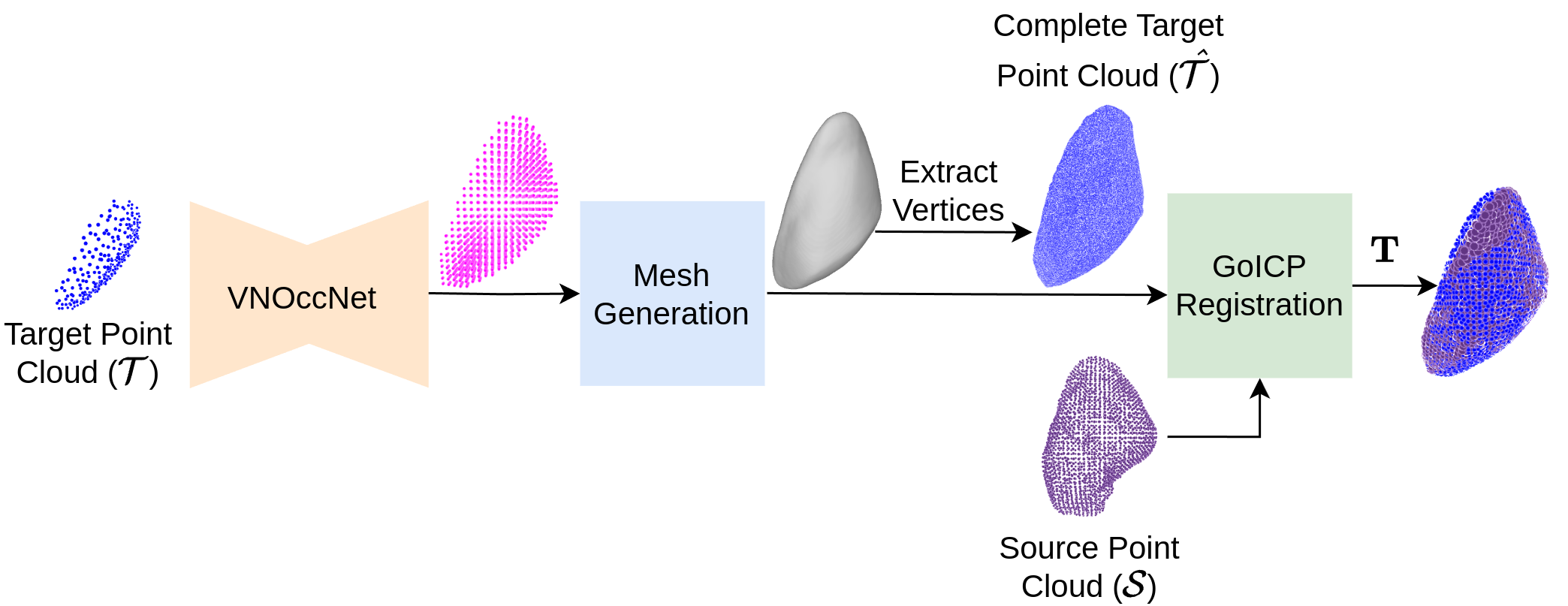}
    \caption{Overview diagram of proposed pipeline: The target point cloud is input to the trained VN-OccNet to output occupancy points (magenta). An integrated mesh generation reconstructs a mesh from the predicted occupancy. The mesh vertices are extracted to represent the complete target point cloud and registered with a source point cloud.}
    \label{fig:pipeline}
\end{figure}

\subsection{Problem Definition}
We define the intra-operative partial liver point cloud as the target point cloud $\mathcal{T} = \{ t_i \}_{i=1}^{N} \in \mathbb{R}^3$  and the pre-operative liver point cloud as the source point cloud $\mathcal{S} = \{ s_i \}_{i=1}^{M} \in \mathbb{R}^3$. Let $f_{\theta}({x_i}, z)$ denote the trained VN-OccNet, where $\theta$ represents the learned parameters, $x_i$ represents the $i^{th}$ query point, and $z$ is the latent representation of $\mathcal{T}$. The solution is to utilize $f_{\theta}(x_i, z)$ to reconstruct a complete estimate of the target point cloud, denoted as $\widehat{\mathcal{T}}$, and subsequently perform registration with $\mathcal{S}$.

\subsection{Proposed Pipeline}
\label{pipeline}
Our proposed pipeline illustrated in Fig. \ref{fig:pipeline} consists of three key stages: (1) Generation of occupancy points from a target point cloud using VN-OccNet, (2) Mesh construction from the predicted occupancy points, and (3) Registration, where the complete target point cloud (obtained by extracting vertices from the generated mesh) is aligned with the source point cloud using the Go-ICP registration method.

\subsubsection{Vector Neuron Occupancy Network (VN-OccNet):}
To accomplish the rotation-equivariant point cloud completion task, VN-OccNet \cite{deng2021vector} leverages the original OccNet architecture \cite{mescheder2019occupancy}, with a vector-based equivariant encoder. The decoder is invariant, where all the operations are non-vector. The target point cloud $\mathcal{T} = \{{t_i}\}_{i=1}^{N} \in \mathbb{R}^3$ is input to the encoder network, and outputs equivariant latent vector-list features ${z}$. The query set $\mathcal{X} = \{{x}_i\}_{i=1}^{K} \in \mathbb{R}^3$, set of points that encloses the liver, and latent feature ${z}$ are input to the decoder network to output the occupancy probabilities between 0 and 1. Points with probability greater than or equal to a threshold $c$ correspond to occupancy points that implicitly represent the liver surface. Since VN-OccNet aims to predict whether or not each point in the query point set corresponds to a liver surface point, binary cross-entropy classification loss is utilized to
learn the parameters $\theta$.

\subsubsection{Mesh Generation:}
Meshes are constructed by first identifying the voxels that intersect the liver surface using the Multiresolution Isosurface Extraction (MISE) algorithm \cite{mescheder2019occupancy}. The algorithm starts with a fixed query set with an initial resolution of $32^3$, consisting of $31^3$ voxels, each containing $8$ points. For all $32^3$ points in the query set, the occupancy probability is determined using the trained VN-OccNet. The points inside the surface are identified by applying a threshold value of $c = 0.4$. Voxels that are entirely inside or outside the surface are separated from the surface-intersecting voxels by checking if at least two adjacent points of a voxel differ in occupancy (i.e., one point is inside and the other is outside). Each such ambiguous voxel that intersects the surface is further subdivided into $8$ subvoxels, generating $19$ additional points. The VN-OccNet is queried again to find the occupancy probability for the newly generated query points. This process is repeated until all ambiguous regions have been subdivided down to the finest voxel grid resolution of $128^3$. Finally, the surface-intersecting voxels obtained are passed to the Marching Cubes algorithm \cite{10.1145/37401.37422} to generate the surface mesh.

 \subsubsection{Rigid Registration using Go-ICP:} To evaluate the effectiveness of incorporating the complete intra-operative target surface generated from using the proposed protocol into the registration, we utilize the Go-ICP registration algorithm \cite{yang2015go} to compare the registration outcomes between using the initial, incomplete target point clouds and the complete intra-operative surface meshes generated using the proposed method.
 
 For the remainder of the manuscript, we refer
to these registrations as registration w/ surface completion (when the complete surface mesh is utilized) and registration w/o surface completion (when only the original partial point cloud is utilized). The registration algorithm yields the transformation ($\mathbf{T}$), which aligns the source point cloud to the target point cloud.

\section{Experiments}

\subsection{Datasets}
The experiments utilize the following datasets: an \textit{in silico} phantom dataset is used to train and test VN-OccNet, and an \textit{in vitro} phantom dataset is used to assess the registration performance.

\subsubsection{\textit{In silico} phantoms:}
We used the undeformed No. 1 synthetic phantom constructed based on a patient-specific CT image according to the methods proposed by Yang {\it et al.}~\cite{10793447} briefly described in further detail in the next sub-section. To simulate deformed models, the deformation simulation pipeline described in~\cite{pfeiffer2020non} is followed, which models the liver as a neo-Hookean hyperelastic material
with a Young’s modulus of 2-5 kPa and a Poisson’s ratio of 0.35. Three
forces of magnitude up to 3N and zero boundary conditions, along with the
material properties, were input to the finite element solver to output deformed liver models. The complete target point clouds are obtained by extracting the mesh vertices. In total, 4,969 deformed models were generated.

Training an occupancy network requires transforming liver models into a representation comprising a liver
point cloud, a set of query points enclosing the liver, and corresponding occupancy values that indicate whether each query point lies inside or outside the
liver surface. We followed the pipeline of Stutz \textit{et al.}~\cite{stutz2020learning} to generate this data format. The generated data was divided into training, validation, and testing sets with a ratio of 8:1:1.

\subsubsection{\textit{In vitro} phantoms:}
\begin{figure}[!th]
\centering
\includegraphics[width=0.9\textwidth]{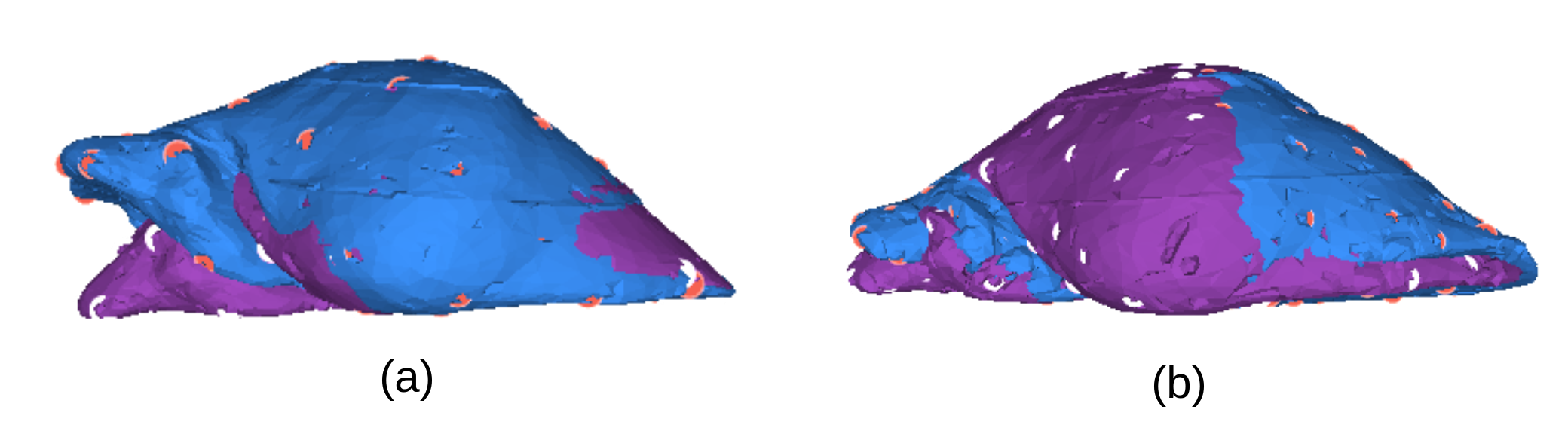}
\caption{The purple phantoms in (a) and (b) represent the undeformed No.1 and phantom No.3, respectively, while the blue phantoms represent their deformed counterparts.
} 
\label{fig:vitroPhantoms}
\end{figure}

We used the \textit{in vitro} phantom dataset described and released in \cite{10793447}, which includes four pairs of underformed and deformed models.  Phantoms were manufactured using synthetic gelatin (Humimic Gelatin \# 0, Humimic Medical, Fort Smith, AK, USA) poured into a 3D-printed mold generated~\cite{merrell2023developing} from a patient-specific liver model obtained from OpenHELP~\cite{kenngott2015openhelp}.  They were deformed by placing wedges of different gradients underneath part of the liver phantom. The surfaces and fiducial marker locations are segmented from CT scans. As Go-ICP requires a long time to estimate rigid registration, we utilized No. 1 and No. 3 phantoms within the dataset in this work, as shown in Fig. \ref{fig:vitroPhantoms}. No. 1 and No. 3 phantoms feature 53 and 176 fiducials, respectively.

\subsection{VN-OccNet Training and Testing Setup}
\label{Target}

\subsubsection{Target Point Cloud Generation:}
\begin{figure}[!ht]
\centering
\includegraphics[width=\textwidth]{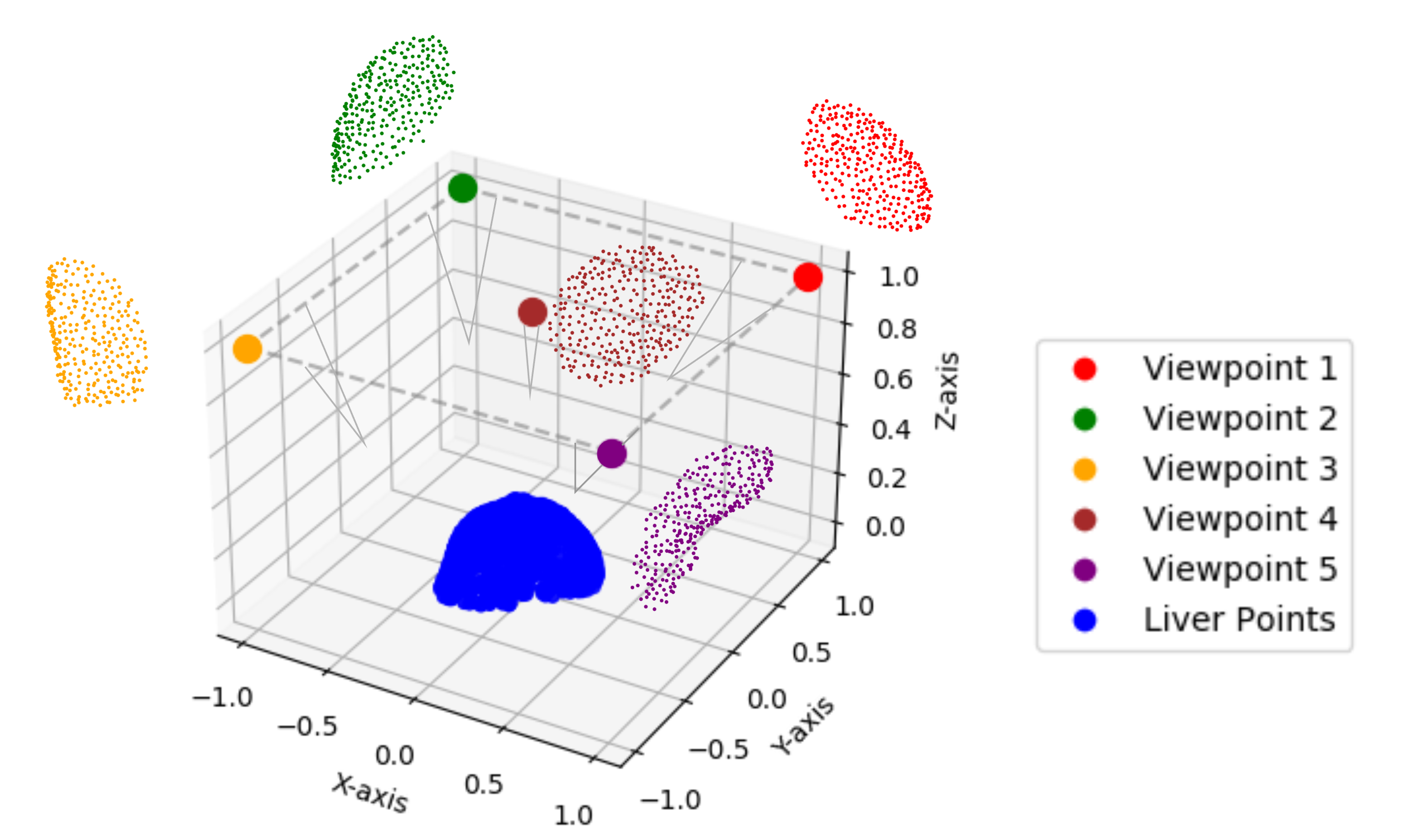}
\caption{Visualization of five target point clouds generated for each intra-operative liver model during testing. Each target is captured from a unique viewpoint, with different regions of the liver obscured to mimic varying intra-operative visibility.
} 
\label{fig: VP}
\end{figure}

In training, target point clouds are dynamically generated from the complete point clouds. First, we crop the posterior surfaces of the deformed liver models. The anterior surface points are then downsampled to $D$ points. A
viewpoint is randomly selected, and the $N$ nearest surface points relative to this viewpoint are extracted to create the target point cloud. For testing, we follow the approach proposed in~\cite{yu2021pointr} to simplify the qualitative analysis, where five distinct viewpoints are selected for
each liver model, as illustrated in Fig.~\ref{fig: VP}.

\subsubsection{Rotation Setup:}
VN-OccNet is trained in three different modes based on the rotation imposed on the target point clouds: without any imposed rotations,
with rotations restricted to the Z-axis, and with SO(3) rotations. We imposed random rotations within the range $[-\frac{\pi}{2}, \frac{\pi}{2}]$ for Z-axis and SO(3) rotation modes. In the non-rotation mode, all target point clouds are aligned and share the same pose. The evaluation involves four train/test rotation settings: I/I, where both training and testing are conducted on non-rotated target point clouds; Z/Z, where both training and testing involve rotations of target point clouds along the Z-axis; Z/SO(3), where training is performed with Z-axis rotations, while testing involves SO(3) rotations; and SO(3)/SO(3), where both training and testing involve SO(3) rotations.

\subsection{VN-OccNet Implementation}
We employed the PyTorch implementation of the  VN-OccNet by Deng {\it et al}. \cite{deng2021vector}. The anterior liver points were downsampled to $D=1000$, and $N=300$ nearest points from the viewpoint were chosen to create the target liver point cloud, yielding a surface visibility of approximately $30\%$. The network was trained with a batch size of $B=8$ and a learning rate of 0.0001, using the Adam optimizer. The optimal threshold value $c$ was determined to be $0.4$ based on the validation dataset. Training was performed for 604 epochs on
a NVIDIA A100 GPU, with all other hyperparameters kept unchanged from the original implementation. To verify the rotation equivariance property, we further compared VN-OccNet with OccNet \cite{mescheder2019occupancy}, which does not include rotation equivariance design.

\subsection{Registration Evaluation Setup}

In a manner similar to the five target point clouds generated for each liver model during testing in the previous subsection~\ref{Target}, five target point clouds are generated for each \textit{in vitro} phantom (No. 1 and No. 3) to perform registration. A set of 100 random SO(3) rotations within the range $[-\frac{\pi}{2}, \frac{\pi}{2}]$ are applied to the target point clouds, resulting in 500 target point clouds at different orientations for each phantom. Registration w/ surface completion is performed by utilizing the complete target point clouds reconstructed from target point clouds featuring a minimum of 30\% visibility. To test the influence of the increment of surface coverage, registration w/o surface completion was performed at three different visibilities of the target point cloud at around $30\%$, $ 40\%$, and $50\%$ by varying the value of $D$, where $D = 1000, 750, \text{and } 600$, respectively.

\subsection{Evaluation Metrics}

 Chamfer Distance (CD-L2), F-Score, and Intersection over Union (IoU) were used to evaluate the performance of complete target surface generation. Moreover, the Target Registration Error (TRE) computed across a set of fiducials was used to assess registration performance.

CD-L2 measures the bidirectional similarity between the complete generated target and ground truth target points:
\begin{equation}
\label{eq:cd}
d_{CD}(\mathcal{\hat{T}}, \mathcal{G}) = \frac{1}{|\mathcal{T}|} \sum_{\mathbf{\hat{t}} \in \mathcal{\hat{T}}} \min_{\mathbf{g} \in \mathcal{G}} \|\mathbf{\hat{t}} - \mathbf{g}\|_2 + \frac{1}{|\mathcal{G}|} \sum_{\mathbf{g} \in \mathcal{G}} \min_{\mathbf{\hat{t}} \in \mathcal{P}} \|\mathbf{g} - \mathbf{\hat{t}}\|_2,
\end{equation}
where $|\mathcal{\hat{T}}|$ and $|\mathcal{G}|$ represent the total number of points in the complete generated target and ground truth target point cloud, respectively, and $d_{CD}(\mathcal{\hat{T}}, \mathcal{G})$ denotes the Chamfer Distance between the point sets $\mathcal{\hat{T}}$ and $\mathcal{G}$.

The F-Score measures the quality of the reconstructed surface by balancing precision $P(d)$ and recall $R(d)$. Precision measures the fraction of complete target points that lie within a threshold $d = 1\,\mathrm{mm}$
 of the nearest ground truth target points, and recall measures the fraction of ground truth target points that lie within $d$ of the nearest complete target points:
\begin{equation}
\text{F-Score}(d) = \frac{2P(d)R(d)}{P(d) + R(d)}.
\end{equation}

IoU  measures the ratio of the intersection to the union of the occupancy values of the generated and ground truth targets, as defined in Eq. \eqref{eq:iou}:
\begin{equation}
\label{eq:iou}
\text{IoU} = \frac{|\text{occ}_{\text{gt}} \cap \text{occ}_{\text{pred}}|}{|\text{occ}_{\text{gt}} \cup \text{occ}_{\text{pred}}|},
\end{equation}
where $\text{occ}_{\text{gt}}$ and $\text{occ}_{\text{pred}}$ represent the ground truth and predicted occupancy values, respectively, and $|\cdot|$ denotes the number of true occupancy labels.

TRE measures the Euclidean distance between the estimated target fiducial markers $\mathbf{F}_{\mathrm{tgt},i}$ and the transformed source fiducial markers $\mathbf{T}(\mathbf{F}_{\mathrm{src},i})$:

\begin{equation}
\label{eq:tre}
 \text{TRE} = \frac{1}{F} \sum_{i=1}^F \|\mathbf{F}_{\mathrm{tgt},i} - \mathbf{T}(\mathbf{F}_{\mathrm{src},i})\|_2,   
\end{equation}

\noindent where $F$ is the total number of fiducial markers.

\section{Results}
\subsection{Mesh Construction Results}
Table \ref{tab: perf} presents the performance of the intra-operative surface completion for OccNet and VN-OccNet across four different train/test configurations: I/I, Z/Z, Z/SO(3), and SO(3)/SO(3). OccNet achieves the best performance across all metrics in the I/I setting. However, its performance significantly deteriorates when tested with target liver point clouds with rotations along the Z-axis and SO(3).

\begin{table}
\caption{Performance of OccNet and VN-OccNet: I/I represents training and testing without any rotations of target point cloud, Z/Z denotes training and testing with rotations along the Z-axis, Z/SO(3) involves training with Z-axis rotations and testing with SO(3) rotations, and SO(3)/SO(3) refers to both training and testing with SO(3) rotations. CD-L2 and IoU represent the Chamfer Distance (in mm) and Intersection over Union.}
\label{tab: perf}
\centering
\setlength{\tabcolsep}{0.8em}
\begin{tabular}{c| c c c c }
\hline
   & I/I & Z/Z & Z/SO(3) & SO(3)/SO(3) \\
\hline
\multicolumn{5}{c}{OccNet} \\
\hline
F-Score  & $ 0.57 \pm 0.20 $& $0.47 \pm 0.17$   &  $ \mathbf{0.07 \pm 0.07} $ &  $0.33 \pm 0.12 $ \\
CD-L2   & $ 2.75 \pm 1.52 $ & $3.51 \pm 1.88$   &   $ \mathbf{29.43 \pm 13.41}$&  $4.87 \pm 2.27 $ \\
IoU  &  $ 0.89 \pm 0.06 $&$0.86 \pm 0.07$  &  $\mathbf{0.35 \pm 0.18 }$ &  $0.81 \pm 0.07$  \\
\hline
\multicolumn{5}{c}{VN-OccNet} \\
\hline
F-Score & $0.52 \pm 0.17$ & $0.51 \pm 0.18$ & $\mathbf{0.51 \pm 0.18}$ & $0.50 \pm 0.18$ \\
CD-L2    &$3.30 \pm 1.91$& $3.29 \pm 1.89$ & $\mathbf{3.25 \pm 1.83}$ & $3.41 \pm 2.02$ \\
IoU  & $0.87 \pm 0.07$ & $0.87 \pm 0.07$  & $\mathbf{0.87 \pm 0.06}$ & $0.86 \pm 0.07$ \\
\hline
\end{tabular}
\end{table}
While data augmentation along SO(3) improves performance to some extent, it remains impractical to encode all possible SO(3) rotations during training, making the model susceptible to failure when encountering unseen liver poses during testing. Although VN-OccNet does not outperform OccNet in the I/I setting, it maintains consistent performance across Z/Z, Z/SO(3), and SO(3)/SO(3), demonstrating robustness to rotations. 
\begin{figure}[!ht]

\includegraphics[width=0.9\textwidth]{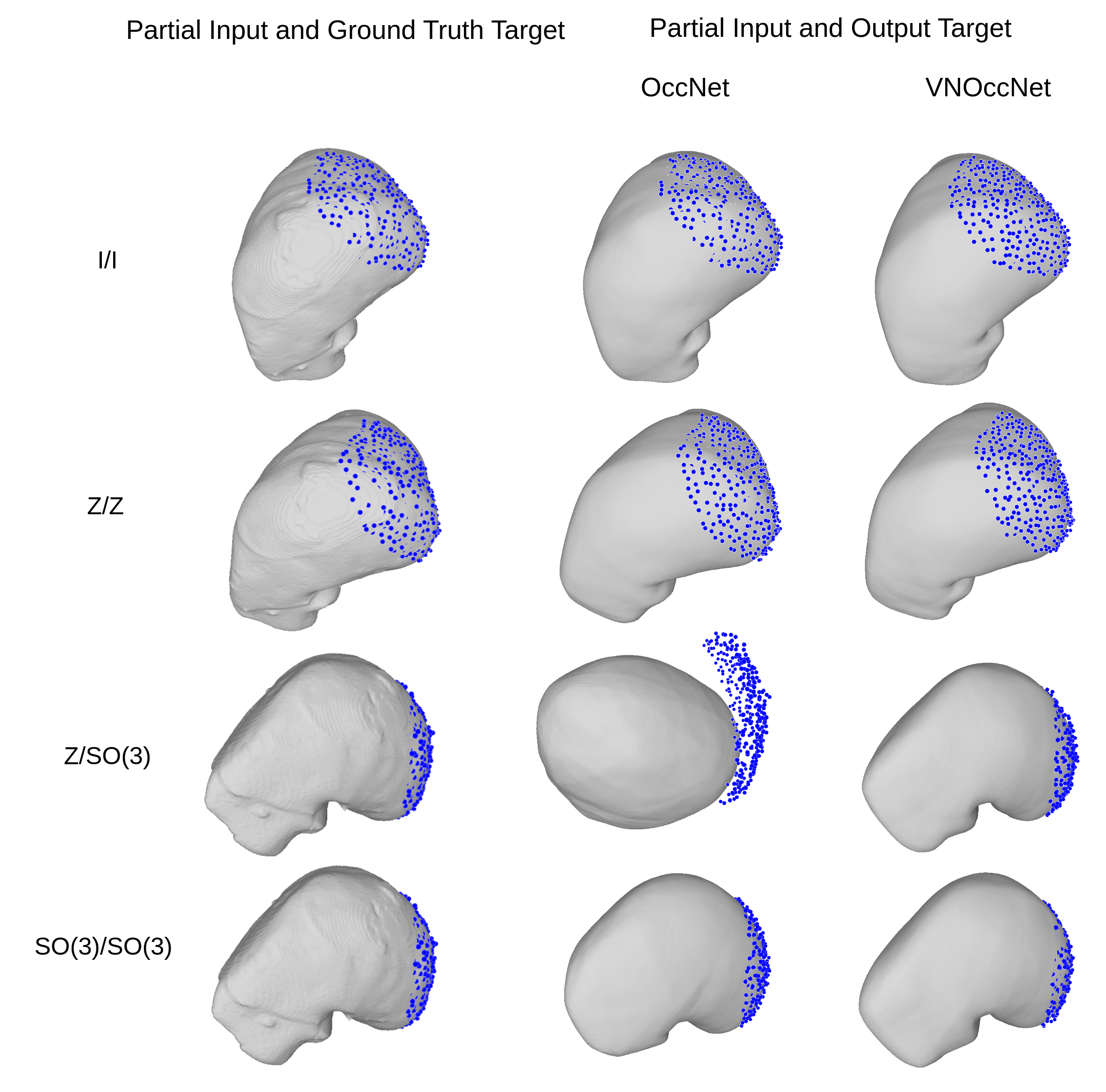}
\caption{Visualization of mesh reconstruction results: The first column displays the target input point clouds alongside the ground truth target mesh. The second column presents the complete target liver surfaces generated by OccNet and VN-OccNet. Qualitative results are shown for four train/test configurations: I/I (first row), where both training and testing involve non-rotated target point clouds; Z/Z (second row), where training and testing are restricted to Z-axis rotations; Z/SO(3) (third row), where training uses Z-axis rotations and testing employs SO(3) rotations; and SO(3)/SO(3) (fourth row), where both training and testing involve SO(3) rotations.}
\label{fig: OPsurface}
\end{figure}

Unlike OccNet, VN-OccNet does not show performance improvement due to data augmentation, indicating that its inherent rotation-equivariant properties reduce reliance on augmented training data. The statistical significance of OccNet and VN-OccNet performance was evaluated by comparing all three evaluation metrics using a Wilcoxon Rank Sum test at a significance level of $\alpha = 0.05$. The test yielded p-values less than $\alpha$ (p < $\alpha$), confirming that the performance differences between VN-OccNet and OccNet across various train/test setups are statistically significant.

Fig. \ref{fig: OPsurface} presents the target point clouds, ground truth target mesh surfaces, and complete target mesh surfaces generated by OccNet and VN-OccNet across four different train/test configurations: I/I, Z/Z, Z/SO(3), and SO(3)/SO(3). The performance gap in surface generation is evident in the Z/SO(3) case, where OccNet struggles to generate a target mesh surface that accurately resembles the ground truth target mesh. In the SO(3)/SO(3) setting, the mesh generated by OccNet lacks the correct geometry, indicating an incorrect reconstruction. In contrast, the mesh generated by VN-OccNet is able to capture correct geometry and better alignment of the target point cloud with the reconstructed mesh.

\subsection{Go-ICP Registration Results}

Table \ref{tab:Regres} summarizes the Target Registration Error (in mm) measured between target fiducial markers and source fiducial markers. The measurements are reported for registration w/ and w/o surface completion. The registration w/o surface completion is performed using the target point clouds at three visibility levels ($30\%$, $40\%$, and $50\%$), and registration w/ surface is performed using the completed surface meshes generated from the 30\% visible target point clouds.

\begin{table}[h]
    \caption{Target Registration Error (TRE) (in mm) measured between target fiducial markers and transformed source fiducial markers for phantoms No.1 and No.3. For registration w/o surface completion, TRE is evaluated at intra-operative visibility levels of $30\%$, $40\%$, and $50\%$. For registration w/ surface completion, the full surfaces are reconstructed from the target point clouds with $30\%$ visibility.}
    \label{tab:Regres}
    \centering
    \setlength{\tabcolsep}{0.8em}
    \renewcommand{\arraystretch}{1.2}

    \begin{tabular}{c | c c c | c}
        \hline
        \multirow{2}{*}{Phantom} & \multicolumn{3}{c|}{W/O surface completion} & \multirow{2}{*}{\thead{W/ surface \\ completion}} \\
        & 30\% Visibility & 40\% Visibility & 50\% Visibility & \\
        \hline
        No. 1 & $33.13 \pm 13.89$ & $30.39 \pm 12.60$ & $26.04 \pm 7.81$ & $\mathbf{5.19 \pm 1.34}$ \\
        No. 3 & $39.87 \pm 15.07$ & $32.79 \pm 14.97$ & $25.87 \pm 10.36$  & $\mathbf{3.35 \pm 0.61}$ \\
        \hline
    \end{tabular}
\end{table}

The registration for phantom No. 1 yields the TRE of $33.13 \pm 13.89$ mm for the $30\%$ visibility of target point clouds. However, when using the complete target point clouds generated from the same $30\%$ visible target, the TRE significantly decreased to $5.19 \pm  1.34$ mm. This notable improvement demonstrates the effectiveness of VN-OccNet's surface completion in enhancing registration performance. Furthermore, as the visibility of the target point clouds increased, TRE progressively decreased, confirming that a lower visibility ratio leads to higher registration errors.

 A Wilcoxon Rank Sum test, conducted at a significance level of $\alpha = 0.05$, showed statistically significant differences (p < $\alpha$) between the three cases of registration w/o surface completion and registration w/ surface completion.

\begin{figure}[ht]
\includegraphics[width=\textwidth]{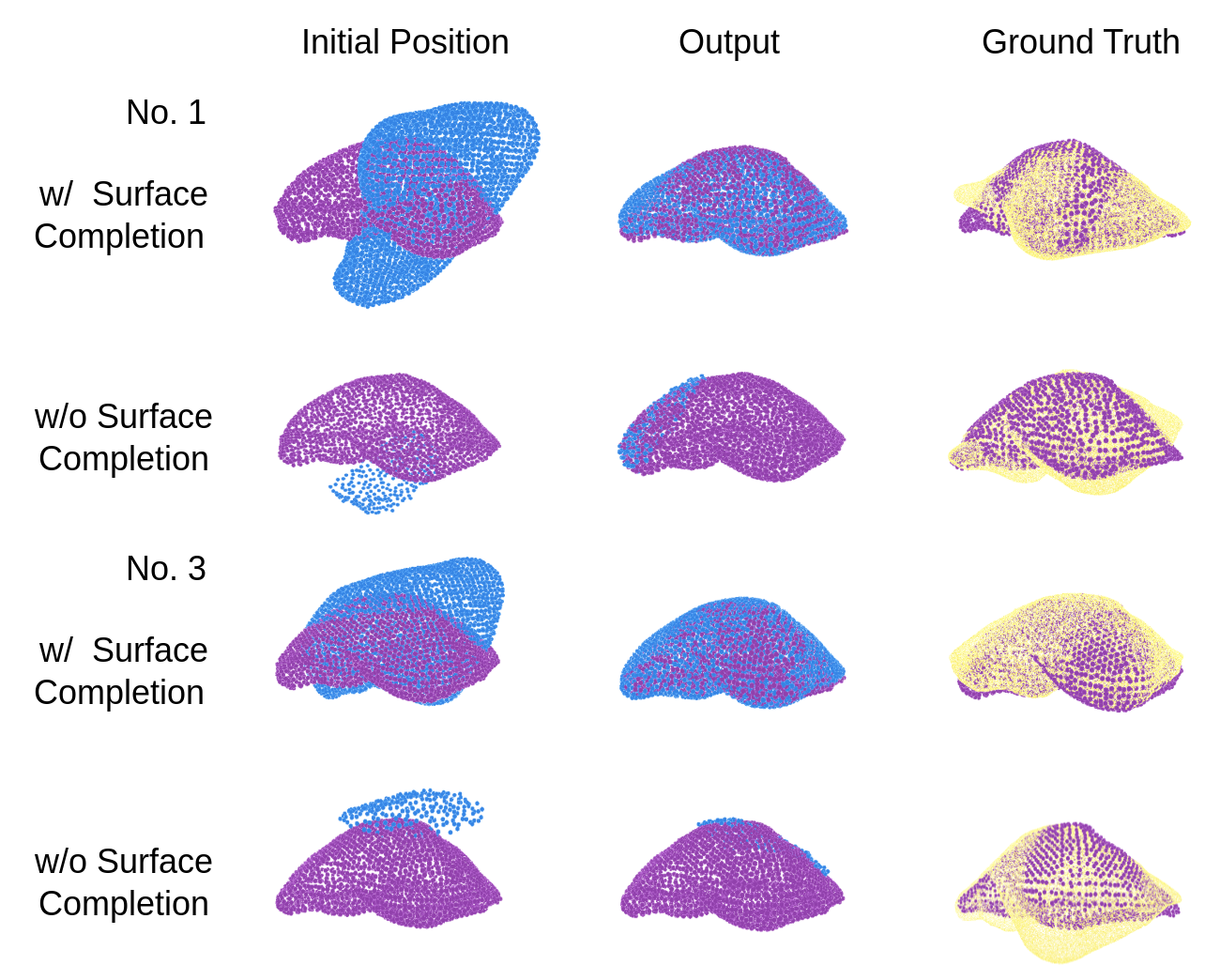}
\centering
\caption{The registration results are displayed for the source point cloud (purple) and the target point cloud (blue) for phantoms No. 1 and No. 3. For easy visualization, the source point cloud is fixed and the target point clouds are moving. The first column represents the initial position of the source and target point clouds. The second column displays the registered results. To see the correctness of the registration, the third column overlays the ground truth target point cloud (yellow). The registrations are performed using partial surface ($30\%$ visibility) and full surface generated from the same partial surface.}

\label{fig: Ipose}
\end{figure}

Fig.~\ref{fig: Ipose} displays the results for registration w/ surface completion and w/o surface completion for \textit{in vitro} phantom No. 1 and No. 3. The registration w/o surface completion struggles to find the correct alignment between the source and target point clouds. The incorrect registration is evident when the complete ground truth target is overlaid onto the registered results. However, registration w/ surface completion is able to correctly align the source and target point clouds. 

\begin{figure}[!ht]
\includegraphics[width=\textwidth]{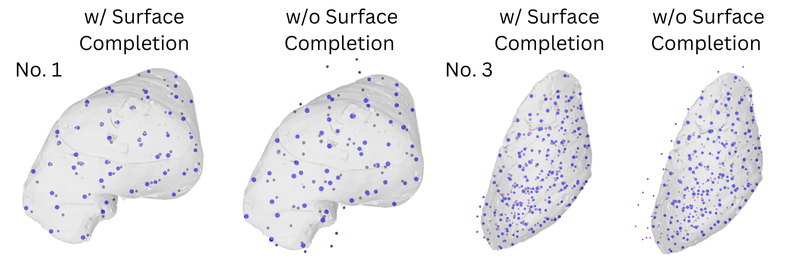}
\centering
\caption{Visualization of target fiducial markers (blue) and registered source fiducial markers (purple) for \textit{in vitro} phantom No 1 and No 3. The target point clouds utilized for registration w/o surface completion have $30\%$ visibility, and the full point clouds are reconstructed from the same targets for registration w/ surface completion. The ground truth target liver surface is overlaid for improved visualization. 
} 
\label{fig: Regres}
\end{figure}

Similarly, Fig.~\ref{fig: Regres} shows the registration between source and target fiducial markers for phantom No. 1 and phantom No. 3. The transformations were determined using a target point cloud with $30\%$ visibility for registration w/o surface completion and a reconstructed completed surface mesh from the same target point cloud for registration w/ surface completion. For the registration w/o surface completion, the registered source fiducial markers are misaligned with respect to the target fiducial markers. In contrast, in registration w/ complete surface, the source and target fiducial markers are closely aligned, demonstrating improved registration.
\section{Discussion}

The partial visibility of intra-operative point clouds poses challenges for both initial rigid and non-rigid registration methods in image-guided liver surgery. To address this issue, we propose a patient-specific, learning-based surface completion approach. In this work, we initially validate the effectiveness of this approach in facilitating the initial rigid registration.

Initial rigid registration has relied on global optimization techniques~\cite{yang2015go} and the establishment of reliable correspondences using handcrafted~\cite{robu2018global} or learning-based methods~\cite{yang2023learning}. However, these approaches are often sensitive to partial liver surfaces and inconsistent point cloud densities. In contrast, our method tackles the problem at its root by employing surface completion through learning-based techniques. Unlike previous completion approaches that require rigid initialization~\cite{jia2021improving} or manual interaction~\cite{foti2020intraoperative}, our method operates without such dependencies. We utilize a vector-based occupancy network—VN-OccNet—due to its robustness to rotation and its ability to generate watertight meshes in just $0.25$ seconds per mesh, making it a strong candidate for this task.

We first verify the rotation-equivariance property of VN-OccNet, which is essential for our application. Recent work~\cite{poudel2025evaluation} evaluated several widely used point completion methods and found them unsuitable for this task. We compared VN-OccNet with the original OccNet, which lacks a rotation-invariant design, under various training and testing conditions, using extensive quantitative and qualitative evaluations. Furthermore, we assessed registration performance by integrating completed surfaces with Go-ICP, a method known for its robustness to partial visibility~\cite{yang2024resolving}. Results show that our reconstructed surfaces significantly reduce registration errors, demonstrating the value of surface completion in improving registration accuracy. However, Go-ICP, which takes approximately 16 seconds per registration, is computationally intensive, suggesting that integrating surface completion directly into learning-based registration methods may offer a more efficient solution.

An additional challenge lies in the variation of liver geometries across different patients, which can hinder surface completion. To overcome this, we adopt a patient-specific training strategy that simulates intra-operative surfaces by deforming a pre-operative patient-specific liver model. This approach enables the network to learn patient-specific geometries and deformation patterns. Our results show that this strategy allows the model to generalize effectively to realistic deformations observed in  \textit{in vitro} phantoms. Nonetheless, the requirement to train a new model for each patient may limit its practicality in resource-constrained environments.

It is also important to note that real intra-operative data may exhibit more complex patterns, such as noise, holes, and occlusions, than those seen in our simulations. This discrepancy poses a challenge for learning-based methods, which generally assume the test data distribution is similar to that of the training data. As such, improving generalization to diverse, realistic intra-operative point clouds remains an open challenge.

\section{Conclusion and Future Work}

This paper presents a pipeline that incorporates VN-OccNet to generate a complete liver surface from partial intra-operative point cloud data, using the reconstructed surface for subsequent registration. We demonstrate the effectiveness of VN-OccNet in producing accurate and complete surface reconstructions, which significantly enhance registration performance. While partial intra-operative data typically hampers the accuracy of most registration methods, our results indicate that the proposed pipeline provides a promising solution to this challenge.

In future work, we plan to investigate the integration of surface completion with non-rigid registration, explore patient-generic surface completion strategies, incorporate completion into end-to-end learning-based registration pipelines, and develop a more robust simulation pipeline capable of generating realistic intra-operative point clouds with noise, holes, and occlusions.

\begin{credits}
\subsubsection{\ackname} We would like to acknowledge the generous support for this work by the National Institutes of Health – National Institute of General Medical Sciences under Award No. R35GM128877 and the National Science Foundation -
Division of Chemical, Bioengineering and Transport Systems under Award No. 2245152. We would like to thank the Research Computing team at Rochester Institute of Technology~\cite{https://doi.org/10.34788/0s3g-qd15} for providing computing resources for this research. We gratefully acknowledge Bidur Khanal for his insightful feedback and encouragement throughout this work.

\end{credits}

 \bibliographystyle{splncs04}
 \bibliography{refs}

\begin{thebibliography}{10}
\providecommand{\url}[1]{\texttt{#1}}
\providecommand{\urlprefix}{URL }
\providecommand{\doi}[1]{https://doi.org/#1}

\bibitem{deng2021vector}
Deng, C., Litany, O., Duan, Y., Poulenard, A., Tagliasacchi, A., Guibas, L.J.: Vector neurons: A general framework for so (3)-equivariant networks. In: Proceedings of the IEEE/CVF International Conference on Computer Vision. pp. 12200--12209 (2021)

\bibitem{foti2020intraoperative}
Foti, S., Koo, B., Dowrick, T., Ramalhinho, J., Allam, M., Davidson, B., Stoyanov, D., Clarkson, M.J.: Intraoperative liver surface completion with graph convolutional vae. In: Uncertainty for Safe Utilization of Machine Learning in Medical Imaging, and Graphs in Biomedical Image Analysis: Second International Workshop, UNSURE 2020, and Third International Workshop, GRAIL 2020, Held in Conjunction with MICCAI 2020, Lima, Peru, October 8, 2020, Proceedings 2. pp. 198--207. Springer (2020)

\bibitem{heiselman2018characterization}
Heiselman, J.S., Clements, L.W., Collins, J.A., Weis, J.A., Simpson, A.L., Geevarghese, S.K., Kingham, T.P., Jarnagin, W.R., Miga, M.I.: Characterization and correction of intraoperative soft tissue deformation in image-guided laparoscopic liver surgery. Journal of medical imaging  \textbf{5}(2),  021203--021203 (2018)

\bibitem{heiselman2024image}
Heiselman, J.S., Collins, J.A., Ringel, M.J., Peter~Kingham, T., Jarnagin, W.R., Miga, M.I.: The image-to-physical liver registration sparse data challenge: comparison of state-of-the-art using a common dataset. Journal of Medical Imaging  \textbf{11}(1),  015001--015001 (2024)

\bibitem{jia2021improving}
Jia, M., Kyan, M.: Improving intraoperative liver registration in image-guided surgery with learning-based reconstruction. In: ICASSP 2021-2021 IEEE International Conference on Acoustics, Speech and Signal Processing (ICASSP). pp. 1230--1234. IEEE (2021)

\bibitem{kenngott2015openhelp}
Kenngott, H.G., W{\"u}nscher, J., Wagner, M., Preukschas, A., Wekerle, A.L., Neher, P., Suwelack, S., Speidel, S., Nickel, F., Oladokun, D., et~al.: Openhelp (heidelberg laparoscopy phantom): development of an open-source surgical evaluation and training tool. Surgical endoscopy  \textbf{29},  3338--3347 (2015)

\bibitem{lin2016video}
Lin, B., Sun, Y., Qian, X., Goldgof, D., Gitlin, R., You, Y.: Video-based {3D} reconstruction, laparoscope localization and deformation recovery for abdominal minimally invasive surgery: a survey. The International Journal of Medical Robotics and Computer Assisted Surgery  \textbf{12}(2),  158--178 (2016)

\bibitem{10.1145/37401.37422}
Lorensen, W.E., Cline, H.E.: Marching cubes: A high resolution {3D} surface construction algorithm. In: Proceedings of the 14th Annual Conference on Computer Graphics and Interactive Techniques. p. 163–169. SIGGRAPH '87, Association for Computing Machinery, New York, NY, USA (1987). \doi{10.1145/37401.37422}, \url{https://doi.org/10.1145/37401.37422}

\bibitem{merrell2023developing}
Merrell, K., Jackson, P., Simon, R., Linte, C.: Developing and evaluating the fidelity of patient specific kidney emulating phantoms for image-guided intervention applications. In: Medical Imaging 2023: Image-Guided Procedures, Robotic Interventions, and Modeling. vol. 12466, pp. 318--323. SPIE (2023)

\bibitem{mescheder2019occupancy}
Mescheder, L., Oechsle, M., Niemeyer, M., Nowozin, S., Geiger, A.: Occupancy networks: Learning {3D} reconstruction in function space. In: Proceedings of the IEEE/CVF conference on computer vision and pattern recognition. pp. 4460--4470 (2019)

\bibitem{pfeiffer2020non}
Pfeiffer, M., Riediger, C., Leger, S., K{\"u}hn, J.P., Seppelt, D., Hoffmann, R.T., Weitz, J., Speidel, S.: Non-rigid volume to surface registration using a data-driven biomechanical model. In: Medical Image Computing and Computer Assisted Intervention--MICCAI 2020: 23rd International Conference, Lima, Peru, October 4--8, 2020, Proceedings, Part IV 23. pp. 724--734. Springer (2020)

\bibitem{poudel2025evaluation}
Poudel, N., Yang, Z., Merrell, K., Simon, R., Linte, C.A.: Evaluation of intraoperative patient-specific methods for point cloud completion for minimally invasive liver interventions. In: Medical Imaging 2025: Image-Guided Procedures, Robotic Interventions, and Modeling. vol. 13408, pp. 376--384. SPIE (2025)

\bibitem{robu2018global}
Robu, M.R., Ramalhinho, J., Thompson, S., Gurusamy, K., Davidson, B., Hawkes, D., Stoyanov, D., Clarkson, M.J.: Global rigid registration of ct to video in laparoscopic liver surgery. International Journal of Computer Assisted Radiology and Surgery  \textbf{13},  947--956 (2018)

\bibitem{https://doi.org/10.34788/0s3g-qd15}
{Rochester Institute of Technology}: Research computing services (2025). \doi{10.34788/0S3G-QD15}, \url{https://www.rit.edu/researchcomputing/}

\bibitem{stutz2020learning}
Stutz, D., Geiger, A.: Learning {3D} shape completion under weak supervision. International Journal of Computer Vision  \textbf{128},  1162--1181 (2020)

\bibitem{yang2015go}
Yang, J., Li, H., Campbell, D., Jia, Y.: {Go-ICP}: A globally optimal solution to {3D} icp point-set registration. IEEE transactions on pattern analysis and machine intelligence  \textbf{38}(11),  2241--2254 (2015)

\bibitem{yang2024resolving}
Yang, Z., Heiselman, J.S., Han, C., Merrell, K., Simon, R., Linte, C., et~al.: Resolving the ambiguity of complete-to-partial point cloud registration for image-guided liver surgery with patches-to-partial matching. arXiv preprint arXiv:2412.19328  (2024)

\bibitem{yang2023learning}
Yang, Z., Simon, R., Linte, C.A.: Learning feature descriptors for pre-and intra-operative point cloud matching for laparoscopic liver registration. International journal of computer assisted radiology and surgery  \textbf{18}(6),  1025--1032 (2023)

\bibitem{10793447}
Yang, Z., Simon, R., Merrell, K., Linte, C.A.: Boundary constraint-free biomechanical model-based surface matching for intraoperative liver deformation correction. IEEE Transactions on Medical Imaging pp.~1--1 (2024). \doi{10.1109/TMI.2024.3515632}

\bibitem{yu2021pointr}
Yu, X., Rao, Y., Wang, Z., Liu, Z., Lu, J., Zhou, J.: Pointr: Diverse point cloud completion with geometry-aware transformers. In: Proceedings of the IEEE/CVF international conference on computer vision. pp. 12498--12507 (2021)

\end{thebibliography}

\end{document}